\documentclass[letterpaper, 10 pt, conference]{ieeeconf}  % Comment this line out if you need a4paper

\IEEEoverridecommandlockouts                              % This command is only needed if 
\usepackage{etoolbox}
\makeatletter
\patchcmd{\@makecaption}
  {\scshape}  {}
  {}
  {}
\makeatother

\usepackage{threeparttable}
\usepackage[table,xcdraw]{xcolor}
\usepackage[backref]{hyperref}
\usepackage{footnote}
\usepackage{mathrsfs}
\usepackage{amsmath}
\usepackage{amssymb}
\usepackage{graphicx}
\usepackage{color}
\usepackage{epsfig} % for postscript graphics files
\usepackage{mathptmx} % assumes new font selection scheme installed
\usepackage{times} % assumes new font selection scheme installed
\usepackage{multirow}
\usepackage{booktabs}
\usepackage{algorithm}
\usepackage{algorithmicx}
\usepackage{CJK}
\usepackage{float}
\usepackage[noend]{algpseudocode}
\usepackage{url}
\UseRawInputEncoding 

\hypersetup{hidelinks}
\title{\LARGE \bf
ROLL: Long-Term Robust LiDAR-based Localization With Temporary Mapping in Changing Environments
}

\author{Bin Peng, Hongle Xie, Weidong Chen$^{*}$ % <-this % stops a space
\thanks{The authors are with Institute of Medical Robotics and Department
of Automation, Shanghai Jiao Tong University, and Key Laboratory of
System Control and Information Processing, Ministry of Education, Shang hai 200240, China. This work is supported by the National Natural Science Foundation of China under Grant U1813206, the National Key R\&D Program of China under Grant 2020YFC2007500, and the Science and Technology Commission of Shanghai Municipality, China under Grant 20DZ2220400. (E-mail: binpeng@sjtu.edu.cn; xiehongle@sjtu.edu.cn; *corresponding author:
wdchen@sjtu.edu.cn)}% <-this % stops a space
\thanks{$^{1}$ \url{https://github.com/HaisenbergPeng/ROLL}}%
}

\begin{document}

\maketitle
\thispagestyle{empty}
\pagestyle{empty}

\begin{abstract}
Long-term scene changes present challenges to localization systems using a pre-built map. This paper presents a LiDAR-based system that can provide robust localization against those challenges. Our method starts with activation of a mapping process temporarily when global matching towards the pre-built map is unreliable. The temporary map will be merged onto the pre-built map for later localization runs once reliable matching is obtained again. 
We further integrate a LiDAR inertial odometry (LIO) to provide motion-compensated LiDAR scans and a reliable initial pose guess for the global matching module. To generate a smooth real-time trajectory for navigation purposes, we fuse poses from odometry and global matching by solving a pose graph optimization problem. We evaluate our localization system with extensive experiments on the NCLT dataset including a variety of changing indoor and outdoor environments, and the results demonstrate a robust and accurate localization performance for over a year. The implementations are open sourced on GitHub$^1$.
\end{abstract}

\section{INTRODUCTION}
Localization is an essential capability for many robotics applications with self-navigation purposes. Multiple techniques can achieve localization but come with limitations. Global Positioning System (GPS) can provide absolute localization in open areas but suffer from multipath effects near or under tall structures. Without such concerns, camera-based localization systems are cheap, accurate, and easy to be deployed, but they suffer from performance deterioration due to illumination changes and scale uncertainty. Light Detection and Ranging (LiDAR) sensors actively emit laser points to perceive 3D environments and thus inherently immune to those problems. With the development of LiDAR technology, LiDAR prices are now acceptable for many commercial applications. Hence we choose LiDAR as our main sensor. 

\begin{figure}[thpb]
      \centering
%      \framebox{\parbox{3in}{		}}
	  \includegraphics[scale=0.5]{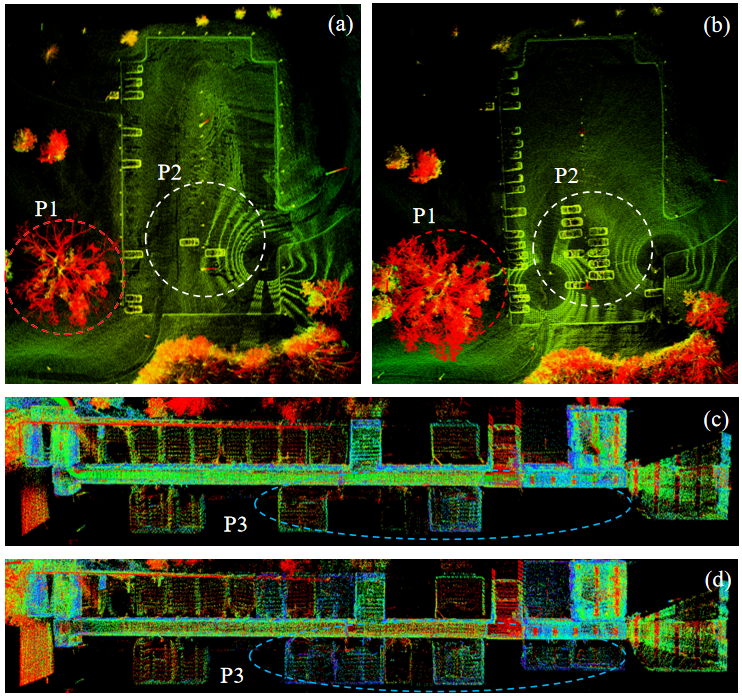}
      \caption{Scene changes in the NCLT dataset. We used the LiDAR data of the same area in different sequences to build a dense local map for comparisons. (a) and (b) are from a parking lot. we can see that trees (P1) are changing their appearances and cars (P2) are parked in different places. (c) and (d) are from a long corridor. Some doors (P3) are opened in (d).}
      \label{scene_change}
\end{figure}

LiDAR-based localization system usually require a dedicated run to build a map of the environment and obtain pose estimation by matching current LiDAR points to the map. For large-scale outdoor scenes, highly compressed formats of LiDAR point cloud maps \cite{ding2020,mGMM2017,poles2021} are employed for localization robustness and storage concerns. However, those maps are not suitable for robots deployed in both indoor and outdoor environments such as smart wheelchairs and field rescue robots. Moreover, constricted and open scenes require different levels of map density for a balance between memory consumption and localization accuracy. The map representation in Posemap \cite{posemap2018} can resolve those concerns. It includes a set of LiDAR keyframes. Each keyframe is observed at a unique location and associated with an observation pose. However, matching towards such a map still suffers from the following issues: (1) In changing environments, poses obtained from global matching towards a pre-built map can be noisy, discontinuous or even erroneous. (2) Global matching is very sensitive to initial values, which are usually provided by matching between consecutive frames. The latter can be seriously affected by fast pitch motion (in direction with sparser LiDAR point density). 

To deal with above issues, a long-term robust LiDAR-based localization system with temporary mapping, titled "ROLL", is proposed. Specifically, a mapping process will be activated temporarily when matching quality is low. Unreliable global matching poses are forfeited in those areas to prevent potential localization failures, and keyframes are registered with poses derived from a LiDAR inertial odometry (LIO). When entering areas with reliable matching, the temporary keyframes will be optimized and then merged into the global map. Temporary mapping allows the system to provide robust localization in significantly changed or even partially unmapped areas. Apart from serving for temporary mapping, LIO can provide motion-compensated LiDAR scans and an accurate pose initials for global matching. And we further combine it with noisy global matching poses in a fusion scheme \cite{qin2019a} to generate smoothed trajectory for navigation purposes. However, the optimization is not guaranteed to reach a global optima and can sometimes gradually go out of bounds. Hence, we implement a consistency check to avoid such a problem. 

The proposed system is evaluated on sequences from a large-scale, long-term open dataset \cite{nclt2016}. The sequences cover a variety of challenging scenarios including vegetation changes, building construction, door opening and closing in a long corridor (shown in Fig. \ref{scene_change}). The system demonstrates a stable and accurate localization performance for over a year.

Our main contributions are summarized as follows:
\begin{itemize} 
\item[$\bullet$] We propose a robust global matching module incorporating temporary mapping, which can prevent localization failures in areas with significant scene changes or insufficient map coverings. The temporary map can be merged onto the global map once matching is reliable again. 
\item[$\bullet$] We extend a fusion scheme to trajectories from LIO and noisy global matching. By implementing a consistency check on the derived odometry drift, we successfully prevent the optimization results from going out of bounds.
\item[$\bullet$] We evaluate the localization system on a large-scale, long-term dataset. The results suggest that the system can survive through a variety of changing environments and exhibit a real-time, robust and accurate localization performance.
\item[$\bullet$] To benefit the community, we open source the implementations.
\end{itemize}

\section{Related Work}
During long-term employment, inconsistencies between changing environment and a pre-built map present challenges for localization systems. In this section we will review several strategies to handle the challenges regardless of sensing modality and discuss their potential applicability to LiDAR localization in a mixed environments. 

In large-scale urban scenes without too many terrain changes, researchers use a selectively compressed representation of the environment as map to encounter scene changes. Early approaches compress registered LiDAR scans into a reflectivity map and achieve centimeter accuracy \cite{levison2007,wolcott2014}. However, the accuracy can be seriously affected in harsh weather when rain or snow changes the road reflectivity. Further studies add structure information in the prior map to increase robustness, such as Gaussian Mixture map \cite{mGMM2017}, combining reflectivity and height information \cite{wan2018,ding2020}, or extracting pole-like landmarks \cite{sefati2017,poles2021}, etc. All those maps above cannot be applied to a mixed environment with both indoor and outdoor scenes. 

Churchill and Newman \cite{churchill2013} accumulate each distinct visual experience of the same place and localizes against those experiences. It is proven to handle some level of environmental changes. Linegar et al. \cite{linegar2015} devised a memory policy to prioritize loading of past experiences. It demonstrates a better performance in long-term localization with less computational resources compared to the previous method. Maddern et al. \cite{maddern2015} further extend experienced-based localization to 2D push-broom LiDARs. It achieves a sub-meter localization accuracy for over a year in changing city scenes. 

Instead of accumulating experiences, some researchers employ mathematical models to predict changes of elementary states in the map. Tilpadi et al. \cite{MC2013} model states of occupancy grids in 2D LiDAR SLAM as Markov processes, achieving a state-of-the-art performance over other existing methods; Krajnik et al. \cite{fremen2017} consider regular feature point in visual SLAM as a combination of harmonic functions; Extending modelling to aperiodic environmental changes, autoregressive moving average model was used in visual inertial SLAM \cite{song2019} and 2D LiDAR SLAM \cite{wang2020}. 

However, both experience-based and prediction-based methods would be computationally expensive for applications in LiDAR localization using point cloud map. Recently, Campos et al. proposed a vision-based SLAM system ORB-SLAM3 \cite{orbslam3}. Rather than localizing against multiple experiences, only one active map is used for global matching and map merging is performed with loop closing. The proposed temporary mapping is similar to their methods. but in our LiDAR system we choose not to maintain multiple non-active maps, because it is both unnecessary and computationally expensive. 

%Integrating a relative localization method into the system is proven to boost long-term robustness. Qin et al. \cite{qin2019a} proposed a general framework for fusing different localization sources such as visual inertial odometry, GPS and barometer etc. Ding et al. \cite{ding2020} and He et al. \cite{he2021} took similar strategies. We extend the fusion method \cite{qin2019a} to fusing LiDAR inertial odometry and feature-map based global matching. However, instead of directly using the odometry drift from last optimization cycle to derive initial values, we implement a consistency check on the odometry drift, which significantly improves robustness to global matching noises.

\section{METHOD}
This section describes the architecture of the proposed localization system in details. The system rests on the following assumptions: (1) Although we only use LiDAR and IMU for long-term localization, there is no limits on the sensors employed to build a globally consistent map. It makes sense because in practical applications a dedicated run with well-equipped sensor suites is usually needed to map the interest site. 
%(2) This paper focuses on dealing with disturbances from scene changes that happen between sessions, which assumes that dynamic changes happening in the session are negligible or considered as outliers. It is a practical assumption for LiDAR sensors that have panoramic views. 
(2) The initial pose of the robot in the pre-built map is known. It can be obtained from GPS or open-sourced place recognition algorithm such as Scan Context \cite{scanContext}.

\subsection{System Overview}
\begin{figure}[thpb]
      \centering
%      \framebox{\parbox{3in}{		}}
	  \includegraphics[scale=0.2]{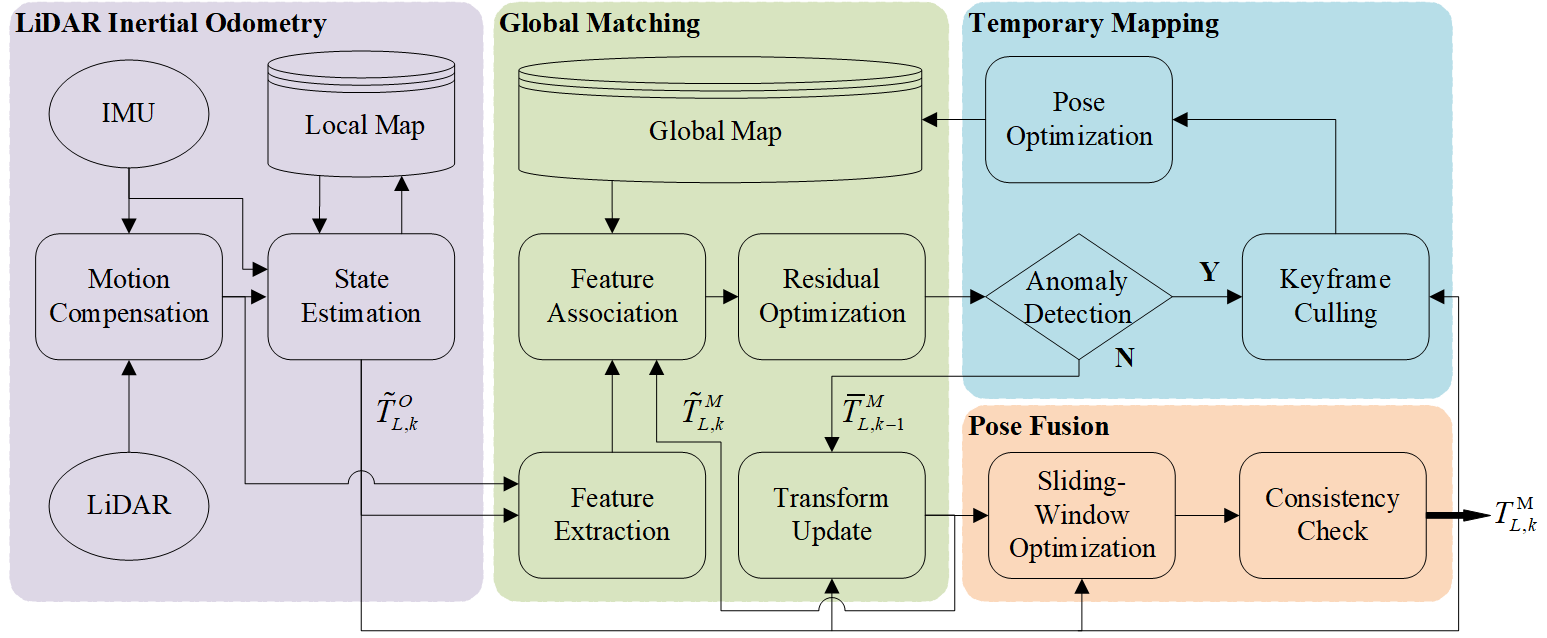}
      \caption{The proposed LiDAR-based localization system architecture.}
      \label{framework}
\end{figure} 

The architecture of our localization system is shown in Fig. \ref{framework}, which consists of four modules: LiDAR inertial odometry (LIO), global matching, temporary mapping, and pose fusion. Our system is general to any LIO since we only need ego-motion estimates and motion-compensated LiDAR scans from the odometry. Global matching module (Section \ref{globalMatching}) projects the extracted LiDAR feature points onto the global map with an initial guess derived from the odometry. Then it performs an optimization over current robot pose to reduce the residuals between the associated feature points. In temporary mapping module (Section \ref{temporaryMapping}), matching inlier ratio is calculated to detect areas with significant scene changes or insufficient map coverings, which we will refer as anomalies in the following. Temporary mapping will be activated upon anomaly detection. To generate a smooth trajectory, odometry poses and global matching poses are fused by solving a graph optimization problem in pose fusion module (Section \ref{sectionFusion}).

%\subsection{Global Map}

\subsection{Global Matching}
\label{globalMatching}
The global matching module is adapted from LOAM \cite{loam2014}, with some enhancements in feature association. A brief introduction is needed for understanding the following chapters. 
\subsubsection{Global Map} 
The global map contains a series of keyframes that are organized with the corresponding optimized observation poses. The incoming motion-compensated LiDAR frames are downsampled, and divided into edge features and surface features based on the smoothness. The features are combined together to form a feature keyframe every time the robot has covered a certain translational or rotational distance. The global map will be updated when temporary mapping is triggered.

Reasons for choosing such feature keyframes instead of a single map are as follows: (1) Feature keyframes are flexible, because the related poses can be optimized and adjusted for loop closure or map extension. (2) In global matching, extracting nearby point clouds is more efficient when using such feature keyframes. (3) It allows for varied density in different map regions. For operation in large-scale sites, the global map needs to be downsampled further for reducing memory consumption. However, for applications that cover both indoor and outdoor scenes, the voxel filter size for downsampling should be varied between confined and open environments. A simple algorithm is proposed here to determine the voxel size. We define the space is confined if ratio of close LiDAR points (distance smaller than $d_c$ in meter) in the feature keyframe there is over a certain threshold ($r_c$). For confined areas such as corridors, the voxel filter size is relatively smaller to increase the coverage intensity.
\subsubsection{Feature Extraction and Association}
Denote LiDAR frame as $L$, map frame $M$ and odom frame $O$, the odom frame is introduced here to describe the inherent drift of the odometry. At time $k$, the motion-compensated LiDAR scan from LIO is downsampled and divided into edge feature points ${\bf E}_k$ and surface feature points ${\bf S}_k$.
For every pose output $\tilde{\bf T}^{O}_{L,k}$ from the LIO. We get the pose guess in the map frame:
\begin{equation}
\label{directTrans}
\tilde{\bf T}^M_{L,k} ={\bf T}^M_{O,k-1}\tilde{{\bf T}}^O_{L,k}
\end{equation}
where ${\bf T}^M_{O,k-1}$ is the transform from the odom frame to the map frame at time $k-1$, here we assume a constant speed model. Notice this pose guess using direct transform is also used for real-time pose output to offer a baseline against the fused poses in Section \ref{sectionFusion}. The comparison is made in Table \ref{ablation2}

With the pose guess $\tilde{{\bf T}}^M_{L,k}$, a kdtree searching within a certain distance is performed on the observation poses of the feature keyframes to extract the nearby clouds for matching. The corresponding feature keyframes are formed together as one local edge map ${\bf P}_k$ and local surface map ${\bf Q}_k$. These local maps are stored in two kdtrees for further fast searching during optimization. 
For every edge point $\bf p_{k,i}\in {\bf E}_k$ after being projected onto the map frame with $\tilde{{\bf T}}^M_{L,k}$, two closest points $ \bf p_{k,i+1}, \bf p_{k,i+2}$ in the local edge map ${\bf P}_k$ are found by kdtree searching. The residual is the point-to-line distance:
\begin{equation}
d_e(\tilde{{\bf T}}^M_{L,k},\bf p_{k,i}, {\bf P}_k) = \frac{\left\| ( \bf p_{k,i} -\bf p_{k,i+1})\times ((\bf p_{k,i} -\bf p_{k,i+2})\right\|}{\left\| \bf p_{k,i+1}-\bf p_{k,i+2}\right\|}
\end{equation}
For every surface point $ \bf q_{k,i} \in {\bf S}_k$ after projection, three closest points $ \bf q_{k,i+1}, \bf q_{k,i+2}, \bf q_{k,i+3}$ in the surface map ${\bf Q}_k$ are found by kdtree searching. The residual is the point-to-surface distance:
\begin{footnotesize}
\begin{equation}
d_s(\tilde{{\bf T}}^M_{L,k}, \bf q_{k,i}, {\bf Q}_k) \\
	= \frac{ \left\|
 (\bf q_{k,i}-\bf q_{k,i+1})^T \\ \left [ (\bf q_{k,i+1} -\bf q_{k,i+2})\times (\bf q_{k,i+1} -\bf q_{k,i+3}) \right ] \right \|
}{\left\| (\bf q_{k,i+1} -\bf q_{k,i+2})\times ((\bf q_{k,i+1} -\bf q_{k,i+3})\right\|}
\end{equation}
\end{footnotesize}
\subsubsection{Residual Computation and Transform Update}
We have a nonlinear function of the residual with respect to the LiDAR pose $\tilde{T}^M_{L,k}$, the feature point $m_{k,i}$ and the local map:
\begin{equation}
f(\tilde{{\bf T}}^M_{L,k}) =  \begin{cases}
 &  d_e(\tilde{{\bf T}}^M_{L,k}, m_{k,i}, {\bf P}_k)  \quad{if} \quad m_{k,i} \in {\bf E}_k \\
 &  d_s(\tilde{{\bf T}}^M_{L,k}, m_{k,i}, {\bf Q}_k)  \quad{if} \quad m_{k,i} \in {\bf S}_k 
\end{cases}
\end{equation}
Stacking the residuals of all feature points, we get a column vector $\mathbf{f}(\tilde{{\bf T}}^M_{L,k})$. Since it is highly nonlinear, the optimized pose $ {\bar{\bf T}^M_{L,k}} $ is obtained through iteratively reducing the norm of residual vector $\mathbf{f}(\tilde{{\bf T}}^M_{L,k})$. Please refer to the original paper for further details. $ {\bar{\bf T}^M_{L,k}} $ is then used to update the transform between odom and map frame ${{\bf T}}^M_{O,k}$ for next global matching.

\begin{figure}[hpb]
      \centering
%      \framebox{\parbox{3in}{		}}
	  \includegraphics[scale=0.4]{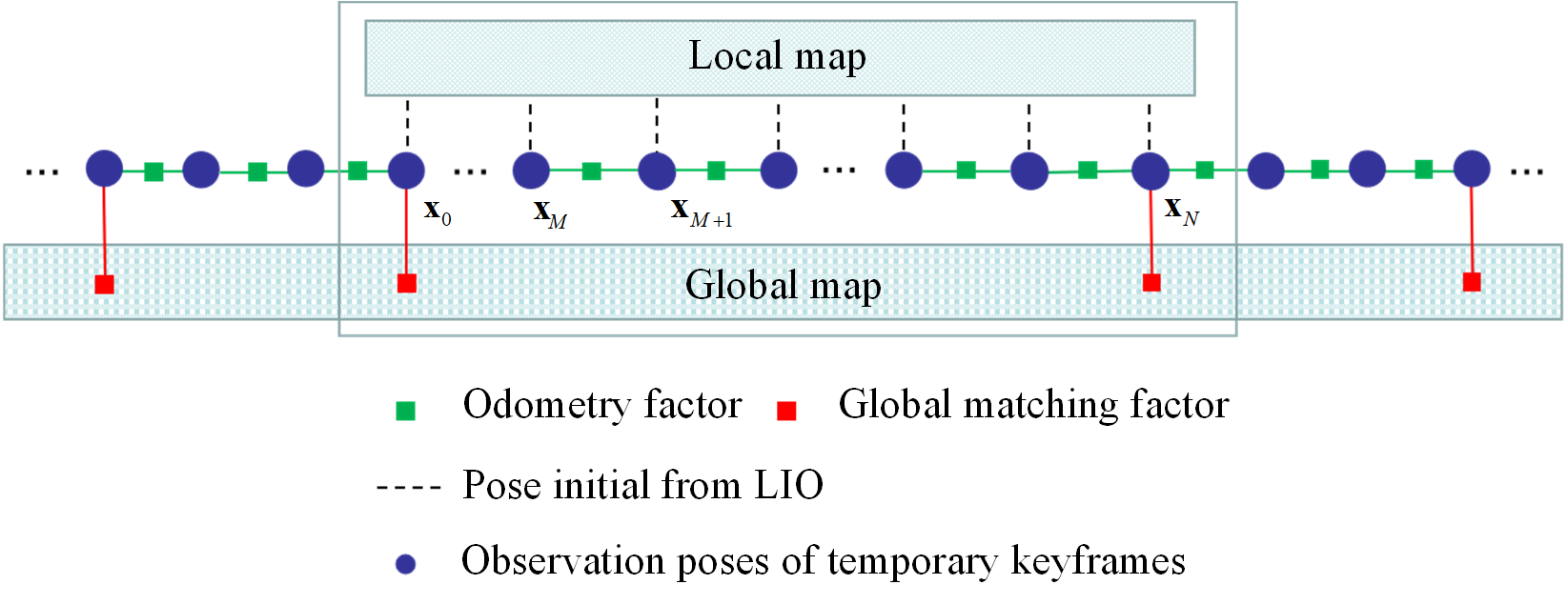}
      \caption{Schematics of the graph optimization on the observation poses of temporary keyframes. Notice the optimization here is different from that of traditional loop closure since we consider observation poses of history keyframes have no need for correction}
      \label{map_merging}
\end{figure} 
\subsection{Temporary Mapping}
\label{temporaryMapping}
In long-term localization, the optimized pose $ {\bar{{\bf T}}^M_{L,k}}$ from global matching is not always suitable for transform update, because map anomalies can render nearby feature keyframes unreliable for matching. 
\subsubsection{Map Anomaly Detection}
For the optimized pose $ {\bar{T}^M_{L,k}}$, we define a criterion called matching inlier ratio to detect those anomalies:
\begin{equation}
\mathbf{\mu}_k = \Gamma(\mathbf{f}(\bar{{\bf T}}^M_{L,k}) ,d_t)
\end{equation}

where $\Gamma(\mathbf{p},q)$ is a function that gets the ratio of elements smaller than $q$ in $\mathbf{p}$. Here $d_t$ is an empirical value, we choose  $d_t = 1.0$, which suffices for all sessions in the experiments. $\mathbf{\mu}_k$ is designed as a rather tolerant way of inspecting the inconsistencies between current sensor observation and the history map so that temporary mapping is activated only when significant inconsistency occurs.
\subsubsection{Keyframe Culling}
When $\mathbf{\mu}_k$ is below a certain threshold $\mathbf{\mu}_{E}$, the system goes into temporary mapping mode. In this mode,  ${\bar{{\bf T}}^M_{L,k}}$ is no longer used for transform update and robot poses transformed from LIO using constant ${{\bf T}}^M_{O,k}$ are used to cull feature keyframes. The culling policy is the same as the one used for building a global map. These keyframes, different from history keyframes, only exists temporarily and thus get the name ``temporary keyframes''. Beware that temporary keyframes are kept as certain number of frames in a sliding-window fashion when no map anomalies is detected. As indicated in Fig. \ref{map_merging}, before starting temporary mapping at pose $x_M$, $M-1$ temporary keyframes composed of recent observations with global matching poses are already stored. It is designed for later pose optimization because global matching poses near $x_M$ is considered unreliable when faced with significant map anomalies. In the mean time, $\mathbf{\mu}_k$ is being calculated for potential map merging with a higher threshold $\mathbf{\mu}_{M}$. Once $\mathbf{\mu}_k$ is above  $\mathbf{\mu}_{M}$, the global matching pose $ {\bar{{\bf T}}^M_{L,k}} $ is considered reliable again to update ${{\bf T}}^M_{O,k}$. 
\subsubsection{Pose Optimization}
To merge the temporary keyframes, a pose graph optimization is performed on the observation poses ${\bf \chi}_t = \{{\bf x}_0, {\bf x}_1,..., {\bf x}_N\}$, where ${\bf x}_i \in SE(3)$. As shown in Fig. \ref{map_merging}. ``Odometry Factor'' represents an association between two poses, the residuals for the optimization can be represented as:
\begin{equation}
\mathbf{r}_{i,j} = {Log(\tilde{\bf x}_{ij}^{-1} {\bf x}_i^{-1} {\bf x}_j)}
\end{equation}
where $\tilde{x}_{ij}$ represents the odometry observation between pose ${x}_i$ and ${x}_j$, the logarithmic operation is mapping from Lie group to Lie algebra $\mathbf{se}(3)$ in its vector form.  ``Global Matching Factors'' are unary factors that defines a residual as:
\begin{equation}
\mathbf{r}_{i} = {Log(\tilde{\bf x}_{i}^{-1} {\bf x}_i)}
\end{equation}
where $\tilde{\bf x}_i$ represents pose constraint from reliable global matching. The residuals are optimized by solving a nonlinear least square problem:
\begin{equation}
\chi^*_t = \displaystyle \min_{ \chi_t} \left\{ \sum_{0 < i < N-1}^{}{{\left\| {\bf r}_{i,i+1}\right\|}_{{\Omega}_i^O}} + {\left\| {\bf r}_{0}\right\|}_{{\Omega}_0^G} +
{\left\| {\bf r}_{N}\right\|}_{{\Omega}_N^G}
 \right\}
\end{equation}
where Mahalanobis norm ${\left\| {\bf r}\right\|}_{{\Omega}} = {\bf r}^T {\Omega}^{-1} {\bf r} $. The covariance matrices ${\Omega}_i^O$ and ${\Omega}_i^G$ are empirically set as constant diagonal matrices. The temporary keyframes after optimization are then merged into the global map by replacing nearby history feature keyframes.

\subsection{Pose Fusion}
\label{sectionFusion}
%% should talk about handling divergent cases!!!
The key for pose fusion is to get an optimized odometry drift (denoted as ${\bf T}^{M}_{O}$) since we output pose estimation as fast as LIO. The real-time pose output for every odometry pose $\tilde{{\bf T}}^O_{L,k}$:
\begin{equation}
\label{fusion}
{\bf T}^M_{L,k} = {\bf T}^M_{O} \tilde{{\bf T}}_{L,k}^O
\end{equation}
\subsubsection{Sliding-window Optimization}
we extend the fusion scheme  \cite{qin2019a} to fuse continuous odometry poses with noisy global matching poses. In essence, for every reliable global matching pose arrives (absence of map anomalies), denote $n$ accumulated odometry poses ${\bf \chi}_O = \{ \tilde{\bf T}^O_{L,k-n+1},\tilde{\bf T}^O_{L,k-n+2},...,\tilde{\bf T}^O_{L,k} \}$ and $m$ global matching poses  ${\bf \chi}_M = \{ \tilde{\bf T}^M_{L,k-n+1}, \tilde{\bf T}^M_{L,k-n+2},...,\tilde{\bf T}^M_{L,k}\}$. A Maximum Likelihood Estimation problem is formulated over the robot poses in the map frame ${\bf \chi}_f = \{{\bf T}_{k-n+1},{\bf T}_{k-n+2},...,{\bf T}_{k}\}$ to derive  ${\bf T}^M_O$, notice here we omit the expression of frames in the notations for simplicity. Notice here we follow the convention used in \cite{qin2019a} by representing poses as ${\bf T} = \{{\bf p}, {\bf q}\}$, where $\bf p$ and $\bf q$ are the corresponding position and orientation quaternion of the pose $\bf T$. Assuming all measurements over robot poses are independent and have Gaussian distributions $p(z_k | {\bf \chi}_f) \sim {\mathcal N}(\tilde{z}_k,{\Omega}_k)$, the problem can be derived as:
\begin{equation}
\begin{aligned}
 {\bf \chi}_f^* =
 \mathop{\arg\min}\limits_{{\bf \chi}_f} \left\{ {\sum_{ i = k-n+1}^{k-1} {\left\| ({z}_i^O - h^O({\bf T}_{i},{\bf T}_{i+1})) \right\| }_{{\Omega}_i^O}} \right. \\
  \left. + \sum_{j = k-m+1}^{k} {\left\| ({z}_j^M - h^M({\bf T}_{j})) \right\| }_{{\Omega}_j^M}  \right\} 
\end{aligned}
\end{equation}
For the odometry pose measurement, the residuals:
\begin{equation}
 {z}_i^O - h^O({\bf T}_{i},{\bf T}_{i+1})
= \begin{bmatrix}
\tilde{{\bf q}}^{O^{-1}}_{L,i}(\tilde{\bf p}^{O}_{L,i+1}-\tilde{\bf p}^{O}_{L,i})
\\
\tilde{\bf q}_{L,i}^{-1} \tilde{{\bf q}}_{L,i+1}
\end{bmatrix}
\ominus 
\begin{bmatrix}
{\bf q}^{{-1}}_{i}({\bf p}_{i+1}-{\bf p}_{i})
 \\
{\bf q}_i^{-1} {\bf q}_{i+1}
\end{bmatrix}
\end{equation}
where symbol $\ominus$ defines an operation that gets the difference between two poses. For translation part, it is element-wise subtraction, while for quaternion  ${\bf q}_i = \{ q_{i,s},{\bf \check{q}}_i\}$, it is element-wise subtraction of the imaginary parts ${\bf \check{q}}_i$. As for the global matching pose measurements, the residuals: 
\begin{equation}
{z}_j^M - h^M({\bf T}_{j}) = \begin{bmatrix}
\tilde{{\bf p}}^{M}_{L,j}
\\
\tilde{{\bf q}}^{M}_{L,j}
\end{bmatrix}
\ominus 
\begin{bmatrix}
{{\bf p}}_j
\\
{{\bf q}}_j
\end{bmatrix}
\end{equation}
To reduce the computation burden, we set $n< n_t$ so that newest odometry pose will pop out the oldest pose in the queue when the queue size is over the limit. The above least square problem can be solved using an open-sourced non-linear solver Ceres \cite{ceres}. Let's denote the optimized robot poses ${\bf \chi}_f^* = \{{\bf T}_{k-n+1}^*, {\bf T}_{k-n+2}^*,..., {\bf T}_{k}^*\}$, and for every ${\bf T}_i^*$, the corresponding odometry drift:
\begin{equation}
{\bf T}^{M^*}_{O,i} = {\bf T}^{*}_{i}{\bf T}^{O^{-1}}_{L,i}=\{{\bf p}^*_{i}-{\bf q}^*_{i} {\bf q}^{O^{-1}}_{L,i}{\bf p}_{L,i}^O,  {\bf q}^*_{i}{\bf q}^{O^{-1}}_{L,i} \}
\end{equation}
where  ${\bf T}^{O^{-1}}_{L,k} = \{ -{\bf q}_{L,k}^{O^{-1}} {\bf p}_{L,k}^O, {\bf q}_{L,k}^{O^{-1}}\}$. 

\subsubsection{Consistency Check}
In the original algorithm, the most recent odometry drift ${\bf T}_{O,k}^{M^*}$ is used to update ${\bf T}_O^M$. Due to noises of global matching introduced by scene changes, the solution may fall into local optima and keeping using the derived odometry drift for initial value update will lead to divergence eventually. We implement a consistency check on the odometry drift with the intuition that the drift should be small within $n_t$ frames. The drift change is defined as:
\begin{equation}
\Delta{\bf T}_O^M = {\bf T}_{O,k-n+1}^{M^{*^{-1}}}{\bf T}_{O,k}^{M^*} 
\end{equation}
where $\Delta{\bf T}_O^M = \{\Delta{\bf p}^M_O,\Delta{\bf q}^M_O\}$. If distance change $\left\| \Delta{\bf p}^M_O \right\| $ is over a certain threshold, the optimization is deemed as falling into local optima. To reset the value, we use the most recent odometry pose and global matching poses to get a odometry drift, which can be expressed as:
\begin{equation}
{\bf T}^M_O = \begin{cases}
 {\bf T}^{M^*}_{O,k} & \text{ if } \left\| \Delta{\bf p}^M_O \right\| \leqslant  p_t \\
 \tilde{\bf T}^{M}_{L,k}\tilde{\bf T}^{O^{-1}}_{L,k}& \text{ if } \left\| \Delta{\bf p}^M_O \right\| > p_t 
\end{cases}
\end{equation}
%The implementation is summarized in Algorithm \ref{fusion}.
Be aware that the fusion is implemented as an independent thread so that the updated odometry drift will not influence the pose guess for global matching.
%\begin{algorithm}[t]
%\caption{Pose fusion}
%\label{fusion}
%\hspace*{0.02in} {\bf Input:} \hspace*{0.02in}
%Odometry poses $\tilde{\bf y}^O_i$, global matching poses $\tilde{\bf y}^M_j$\\
%\hspace*{0.02in} {\bf Output:} 
%Odometry drift ${\bf y}_{O}^M$, real-time fusion pose ${\bf y}_i$
%\begin{algorithmic}[1]
%\For{ $\tilde{\bf y}^O_i$} 
%\State Generate real-time pose ${\bf y}_i = {\bf y}^M_O \tilde{\bf y}^O_i$
%\State Accumulate odometry poses in ${\bf \chi}_O$
%\EndFor
%
%\For{condition} % For ??????EndFor??
%??\State ...
%??\If{condition} % If ??????EndIf??
%????\State ...
%??\Else
%????\State ...
%??\EndIf
%\EndFor
%
%\While{condition} % While??????EndWhile??
%\State ...
%\EndWhile
%\State \Return result
%\end{algorithmic}
%\end{algorithm}

\begin{figure}%[thpb]
      \centering
%      \framebox{\parbox{3in}{		}}
	  \includegraphics[scale=0.32]{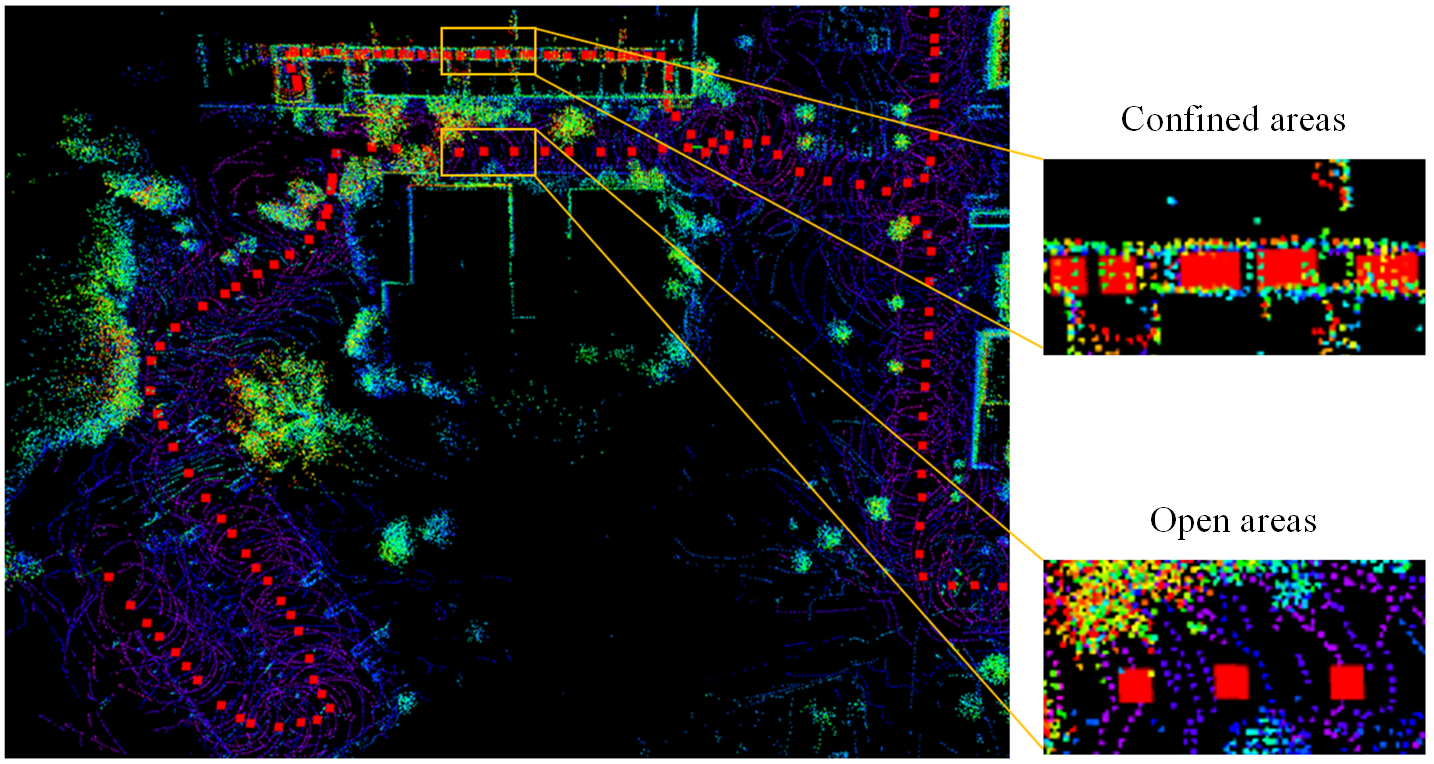}
      \caption{A demonstration of the map used in the localization system. The map is processed with different filter size in open and confined areas, each red dot represents one feature keyframe. The two figures in the right are enlarged views of the areas marked by yellow rectangles, denser red dots can be observed in the upper right of the figure.}
      \label{feature_map}
\end{figure}

\begin{table}[]
\centering
\caption{Localization performance of the proposed algorithm evaluated on the NCLT dataset. }
\label{results}
\begin{threeparttable}
%\begin{tabular}{c|c|c|c|c|c|c|c}
%\hline
%Sequence &
%Length (km)&
%{RMSE (m)} &
%{MAX (m)} &
%{$\Gamma(e_t,0.1) (\%)$} &
%{$\Gamma(e_t,0.2) (\%)$} &
%{$\Gamma(e_t,0.5) (\%)$} &
%{TM Number} \\ \hline
%12-02-02 & 6.2 & 0.189 & 1.481 & 35.029 & 77.503 & 98.440 & 1 \\ \hline
%12-03-17 & 1.4 & 0.180 & 1.171 & 33.108 & 77.858 & 99.052 & 2 \\ \hline
%12-04-29 & 3.1 & 0.226 & 1.474 & 29.352 & 74.760 & 97.292 & 1 \\ \hline
%12-05-11 & 6.0   & 0.248 & 1.390 & 27.670 & 70.706 & 94.833 & 3 \\ \hline
%12-06-15 & 6.3 & 0.260 & 1.777 & 26.330 & 68.132 & 92.597 & 3 \\ \hline
%12-08-04 & 4.1 & 0.213 & 1.196 & 24.919 & 69.633 & 97.578 & 1 \\ \hline
%12-11-17 & 5.5 & 0.286 & 1.142 & 24.553 & 62.379 & 92.812 & 2 \\ \hline
%13-01-10 & 1.1 & 0.277 & 1.049 & 21.203 & 62.457 & 93.631 & 1 \\ \hline
%13-02-23 & 5.2 & 0.281 & 1.153 & 34.250 & 72.793 & 91.576 & 1 \\ \hline

\begin{tabular}{c|c|c|c}
\hline
Sequence &
{RMSE/Max (m)} &
$<0.1/0.2/0.5m (\%)$ &
{TM Number} \\ \hline
12-02-02 & 0.189/1.481 & 35.029/77.503/98.440 & 1 \\ \hline
12-03-17  & 0.180/1.171 & 33.108/77.858/99.052 & 2 \\ \hline
12-04-29  & 0.226/1.474 & 29.352/74.760/97.292 & 1 \\ \hline
12-05-11    & 0.248/1.390 & 27.670/70.706/94.833 & 3 \\ \hline
12-06-15  & 0.260/1.777 & 26.330/68.132/92.597 & 3 \\ \hline
12-08-04  & 0.213/1.196 & 24.919/69.633/97.578 & 1 \\ \hline
12-11-17 & 0.286/1.142 & 24.553/62.379/92.812 & 2 \\ \hline
13-01-10  & 0.277/1.049 & 21.203/62.457/93.631 & 1 \\ \hline
13-02-23  & 0.281/1.153 & 34.250/72.793/91.576 & 1 \\ \hline

\end{tabular}
\end{threeparttable}
\end{table}

\begin{figure}[hpb]
      \centering
%      \framebox{\parbox{3in}{		}}
	  \includegraphics[scale=0.30]{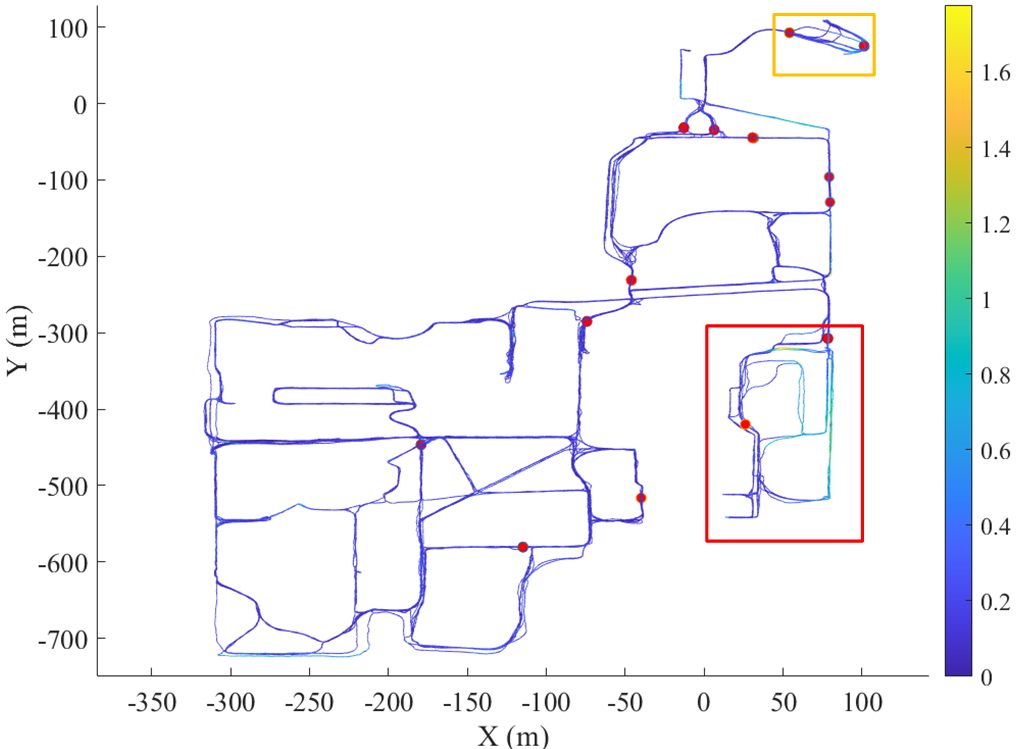}
      \caption{Localization trajectories from the tested nine NCLT sequences colored by translational distance error. The red dots indicate the places where map anomalies were detected. Two types of map anomalies are marked by yellow (significant scene changes) and red (insufficient mapping coverings) rectangles}
      \label{error_traj}
\end{figure} 

\begin{footnotesize}
\begin{table}[]
\centering
\caption{Comparisons with the baseline algorithms}
\label{comp} % it needed to be after caption
\begin{threeparttable}
\begin{tabular}{c|c|c|c}
\hline
Algorithm                 & S.R.(\%) &  RMSE/Max(m) &  $<0.5m (\%)$  \\ \hline
LOAM(M) \cite{loam2014}    &	1.575    &	3.595/7.577		 & 11.236 				 \\ \hline
LOAM(M)+TM                &   72.188  & 4.105/18.738            &39.869            \\ \hline
ROLL 					 	&   \textbf{98.851}   &\textbf{0.248}/\textbf{1.390}             &  \textbf{94.833}           \\ \hline
\end{tabular}
\end{threeparttable}
\end{table}
\end{footnotesize}

\section{EXPERIMENTS}
\label{exp}
All experiments were performed on a desktop with an Intel Xeon W-2102 CPU@2.9 GHz and 16 GB of RAM. The algorithm is tested on ten sequences from the NCLT dataset \cite{nclt2016}. 
The NCLT dataset is a large-scale, multi-modal, long-term anatomy dataset collected by a UGV (Unmanned Ground Vehicles) in the University of North Michigan's North Campus for over a year. It covers both indoor and outdoor environments with a wide variety of scene changes such as snowing, tree leaves falling, building constructions, door opening and closing in long corridors, etc. For example, the weather is snow at day ``12-01-15'', ``13-01-10'', ``13-02-23'' (For simplicity, the date of sequence recording ``year-month-day'' is used to refer a sequence). Further, sequences ``12-05-11'' and ``12-11-17'' covered some unexplored areas, that would require the localization system to be able to extend the pre-built map. The ground truth of the dataset is obtained from a large offline SLAM solution of RTK (Real-time kinematics) GPS and LiDAR scan matching of all the sessions. 

Since this algorithm is mainly designed for localization, we used the ground truth poses of the UGV and LiDAR data recorded at day ``12-01-15'' to build an initial global map and the other nine sequences for localization experiments.  
After applying downsampling on the initial global map, the map size is 53.8 MB for the whole campus with the size of around 800$\times$600 $m^2$. It ends up in 61.9 MB after finishing the last sessions.  A demonstration of downsampling is shown in Figure \ref{feature_map}. We can see that different keyframe density is achieved. 
We choose FAST-LIO2 \cite{fastlio2021} as our LiDAR inertial odometry, since it demonstrates a state-of-the-art performance over other existing methods. %It employs an iterated kalman filter to estimate the ego-motion. Specifally, for every IMU input, it uses a forward propagation to update the states that include pose, speed, angular rates, IMU biases and LiDAR-IMU extrensics; For every LiDAR scan, it applies backward propagation on raw LiDAR points to compensate for motion distortion and then updates the states iteratively to obtain the optimal estimates. LiDAR scans with the optimal poses are registered in a local map organized into an ikd-Tree. Only map points within a certain local region is maintained in the ikd-Tree. 

For localization runs, the initial pose of the UGV is given by the ground truth. After initialization, we directly use the raw data from a Velodyne HDL-32E LiDAR and a Microstrain 3DM-GX3-45 IMU. Some parameters are empirically set as follows: $d_c = 5.0, r_c = 0.3, {\mu}_E = 0.3, {\mu}_M = 0.5, p_t = 0.5, n_t = 100$.

\begin{figure}%[hpb]
      \centering
	  \includegraphics[scale=0.38]{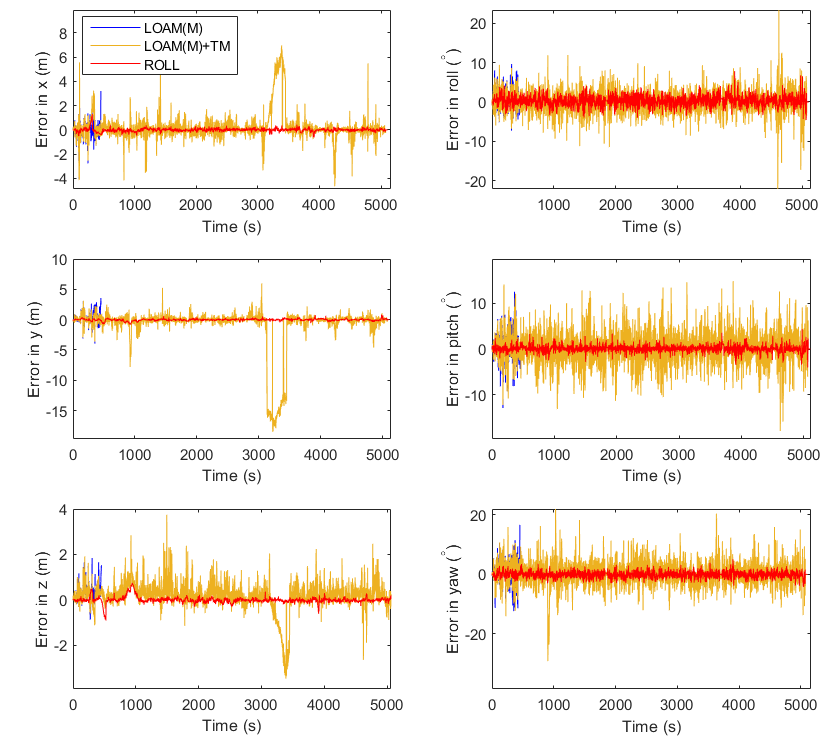}
      \caption{Position and orientation error comparisons with the baseline algorithms }
      \label{XYZerr}
\end{figure} 

\subsection{Long-Term Localization Performance}
As shown in Table \ref{results}, we evaluated the localization accuracy with a commonly used absolute trajectory error \cite{sturm2012}. Specifically, we compared the localization poses ${\bf X}_i ({\bf X}_i \in \mathbf{SE}(3), i=1,...,n)$ with ground truth ${\bf G}_{i} ({\bf G}_i \in \mathbf{SE}(3),i=1,...,m)$ to get a time-dependent absolute translation error ${\bf F}_i $, time synchronization is through a kdtree searching in the timeline of the ground truth within 50 ms. Apart from the RMSE, to present the whole picture of long-term localization performance, we introduce error percentages of the translational distance error ${\bf e}_i = \left\| trans({\bf F}_i) \right \|$. The error percentage is defined as the ratio of ${\bf e}_i$ smaller than a certain threshold among all ${\bf e}_i$. Throughout all sessions, the algorithm demonstrates a RMSE of 0.239 meter and a real-time performance (9.91 Hz, approximately the rate of LiDAR data). 

Temporary mapping is only activated a few times for every session. Places of activations are marked in the error trajectory, as shown in Fig. \ref{error_traj}. Two examples of map anomalies are marked by yellow and red rectangles. From the error trajectory, we can observe that the area marked by a red rectangle exhibits larger localization error relative to other areas. It can be explained by the map extensions in that area. The area marked by a yellow rectangle is a parking lot referred in Fig. \ref{scene_change}. The area is very open with few stable features available for matching, prompting the activations of temporary mapping.

\textbf{Comparisons}. In Table \ref{comp}, we compared our algorithm with the variants of the state-of-the-art LiDAR SLAM LOAM \cite{loam2014} on sequence ``12-05-11''. Specifically, for ``LOAM(M)" we use the same map representation as ROLL for map saving and loading, but we only match towards a pre-built map without point cloud registration. For ``LOAM(M)+TM" we integrate temporary mapping into ``LOAM(M)'' to enable map udpates for scene changes. For a fair comparison, these two baseline algorithms use ground truth maps too, and the localization poses used for comparisons are generated from a direct transform between LiDAR odometry and mapping, similar to Equation \ref{directTrans}. Because not all algorithms can survive through the whole session, we define a success ratio (S.R.) to refer the ratio of frames with ${\bf e}_i$ smaller than one meter among all LiDAR frames. Thanks to temporary mapping, ``LOAM(M)+TM" shows a significant improvement on robustness over ``LOAM(M)''. Aided by LIO and fusion algorithm, ROLL outperforms both of them. The position error comparisons are shown in  Fig \ref{XYZerr}. ``LOAM(M)" failed at around 440 sec due to entering partially unmapped areas; Both ROLL and ``LOAM(M)+TM'' can survive through the whole session, but ROLL exhibits obvious smaller errors than ``LOAM(M)+TM''.

\begin{footnotesize}
\begin{table}[]
\centering
\caption{Ablation study on temporary mapping (TM) and consistency check (CC). The evaluation criterion is success ratio in percentage}
\label{ablation1} % it needed to be after caption
\begin{threeparttable}
\begin{tabular}{c|c|c|c|c}
\hline
Algorithm & 12-02-02 & 12-03-17 & 12-04-29 & 12-05-11                  \\ \hline
ROLL w.o. CC & 6.230           & 7.718           & 9.607           & 0.708           \\ \hline
ROLL w.o. TM & 1.567           & 99.006           & 98.424           & 8.397          \\ \hline
ROLL         & \textbf{98.534} & \textbf{99.107} & \textbf{98.439} & \textbf{98.851} \\ \hline
\end{tabular}
\end{threeparttable}
\end{table}
\end{footnotesize}

\begin{footnotesize}
\begin{table}[]
\centering
\caption{Pose output comparison between pose fusion and direct transform. }
\label{ablation2} % it needed to be after caption
\begin{threeparttable}
\begin{tabular}{c|c|c}
\hline
Method					& RMSE/Max(m)  			 &$<0.1/0.2/0.5m (\%)$                	 \\ \hline
Direct Transform        & 0.240/\textbf{1.751}    & 28.450/70.643/\textbf{95.552}          \\ \hline
Fusion 			 		& \textbf{0.239}/1.777   & \textbf{29.390}/\textbf{71.481}/95.536 \\ \hline
\end{tabular}
\end{threeparttable}
\end{table}
\end{footnotesize}
\begin{figure}%[hpb]
      \centering
%      \framebox{\parbox{3in}{		}}
	  \includegraphics[scale=0.42]{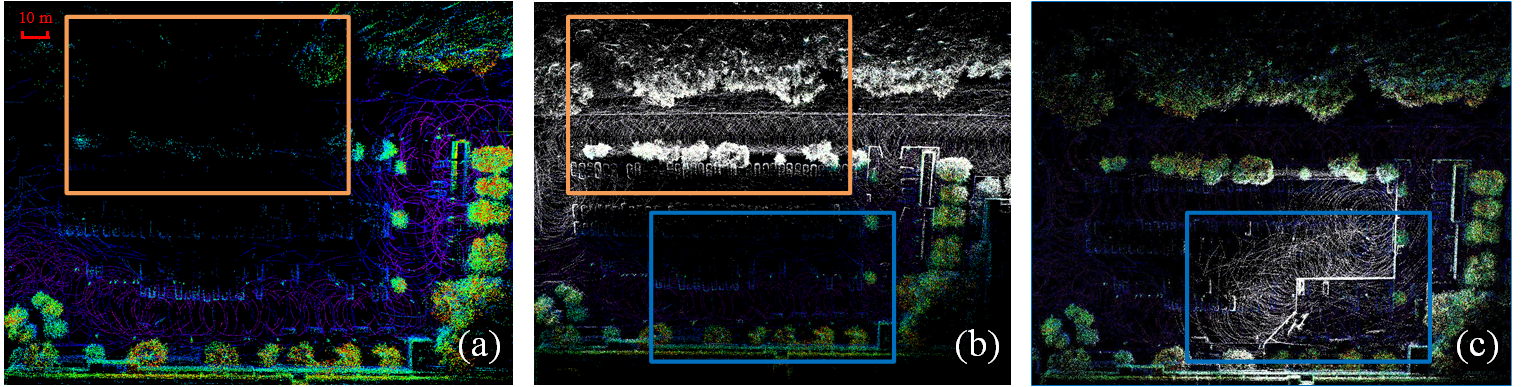}
      \caption{Demonstrations of map updates. Notice the circled areas. (a) The aerial view of the point cloud map in the partially unmapped areas before day ``12-05-11''; (b) The same view as (a) after map update at day ``12-05-11'', the white colored point cloud is the merged temporary map; (c) The same view as (b) after map update at day ``12-06-15'', we can observe that the construction fences are updated onto the global map.}
      \label{TM}
\end{figure}
\subsection{Map Update Performance}
%The algorithm is designed to only update the global map when map anomalies are detected, hence map update is determined by the values of  $\mathbf{\mu}_{E}$ and  $\mathbf{\mu}_{M}$. Before conducting localization tests, we run the algorithm tentatively on a sequence (``12-03-17") with $\mathbf{\mu}_{E}$ set to be 0. It means temporary mapping mode will never be activated. Then the distribution of the matching inlier ratio can be obtained as shown in Fig. \ref{TM}(a). Based on the distribution, $\mathbf{\mu}_{E}$ is set to 0.3 for all sessions. For the threshold $\mathbf{\mu}_{M}$, we set it to 0.5. It is high enough to provide an accurate global matching to correct odometry drift but not too high to never exit temporary mapping. With those thresholds, the algorithm can survive through all nine sequences in the NCLT dataset. 

Examples of map updates can be demonstrated in Fig. \ref{TM}. As shown in Fig. \ref{TM}(a), part of the global map is unexplored before running sequence ``12-05-11'', but the algorithm is able to identify this anomaly and update the map of this region, as indicated by the white point cloud in Fig. \ref{TM}(b). Later in sequence ``12-06-15'', a construction site is also successfully updated onto the global map, as shown in Fig. \ref{TM}(c). All above demonstrate the robustness of the algorithm against map anomalies including insufficient map coverings and significant scene changes.

\subsection{Ablation Study}
To evaluate the contributions of the proposed methods for the whole algorithm, we carried out two sets of ablation experiments. The first set of experiments are performed on four sequences to demonstrate the necessity of temporary mapping and consistency check, as shown in Table \ref{ablation1}. Success rate is chosen as the criterion for evaluation. Without temporary mapping, the algorithm sometimes fails when visiting significantly changed areas in sequence ``12-02-02'' or partially unmapped areas in sequence ``12-05-11''. 
%However, the failure is not a deterministic event due to the stochastic nature of the global matching.
Further, the ablation experiments also clearly explain the necessity of consistency check. The original fusion method, if used without modification, cannot handle even ten percent of each session.

The second set of experiments are performed on all sequences to compare the pose fusion (Equation \ref{fusion}) with the direct transform (Equation \ref{directTrans}), as shown in Table \ref{ablation2}. With a slight sacrifice in the maximum error, pose fusion demonstrates better error percentages overall.

\section{CONCLUSIONS}
We have presented a long-term robust LiDAR-based localization system. It adopts a temporary mapping module to avoid potential localization failures caused by erroneous global matching in significantly changed or partially unmapped areas. Activation of temporary mapping can update those areas on the global map for later localization sessions. Further, the incorporation of a LiDAR inertial odometry provides global matching module with motion-compensated LiDAR points and pose initials unaffected by scene changes. The fusion scheme between global matching and the odometry is improved with a consistency check on the odometry drift after the optimization, which significantly increases robustness of the fusion algorithm to scene changes. The evaluation on the NCLT dataset proves that the proposed system can provide a real-time, accurate and robust localization in changing indoor and outdoor environments for over a  year. In future work, we plan to devise an adaptive scheme on when to entering or exiting temporary mapping for different environments instead of setting constant thresholds. 
%\addtolength{\textheight}{-1cm}   % This command serves to balance the column lengths

%\newpage
\bibliographystyle{ieeetr}
\bibliography{TMM.bib}
\end{document}